\documentclass[10pt]{article}
\usepackage{spconf,amsmath,graphicx}
\usepackage[T1]{fontenc}
\usepackage{booktabs}
\usepackage{multirow}
\usepackage{fancyhdr}
\usepackage{ifthen}
\usepackage{lipsum}

\title{Knowledge-aware Bayesian Co-attention for Multimodal Emotion Recognition}
\name{Zihan Zhao$^1$, Yu Wang$^1{^,}{^2}{^*}$, Yanfeng Wang$^1{^,}{^2}{^*}$\thanks{* Corresponding authors}\thanks{This work was supported by National Key R\&D Program of China (No.2022ZD0162100), National Natural Science Foundation of China (No.62106140), Shanghai Science and Technology Committee (No.21511101100) and Shanghai Key Lab of Digital Media Processing and Transmission (No.18DZ2270700)}}
\address{$^1$Cooperative Medianet Innovation Center, Shanghai Jiao Tong University\enspace$^2$Shanghai AI Laboratory\\
    \small{
        \{zihanzhao, yuwangsjtu, wangyanfeng\}@sjtu.edu.cn
    }}

\begin{document}
\maketitle
\thispagestyle{fancy}%
\begin{abstract}
\vspace{-0.1cm}
Multimodal emotion recognition is a challenging research area that aims to fuse different modalities to predict human emotion. However, most existing models that are based on attention mechanisms have difficulty in learning emotionally relevant parts on their own. To solve this problem, we propose to incorporate external emotion-related knowledge in the co-attention based fusion of pre-trained models. To effectively incorporate this knowledge, we enhance the co-attention model with a Bayesian attention module (BAM) where a prior distribution is estimated using the emotion-related knowledge. Experimental results on the IEMOCAP dataset show that the proposed approach can outperform several state-of-the-art approaches by at least 0.7\% unweighted accuracy (UA).
\end{abstract}
\vspace{-0.3cm}
\section{Introduction}
\vspace{-0.3cm}
\label{sec:intro}
Emotion recognition is the process of identifying human emotion. Multi-modality for emotion recognition has recently received great interest because different modalities can provide complementary clues and the fusion of them can effectively improve the performance \cite{zhao2022multi,9746598,chen20b_interspeech}. For example, the meanings of words and their relation express one's emotion in text modality, while intonation and pitch can also be useful for recognizing emotions in speech modality. In this paper, we focus our research on text and speech modalities.

Emotion datasets usually face the problem of insufficient data \cite{pepino21_interspeech}, mainly due to the difficulties caused by the subjective nature of labeling emotions. The common solution to solve this problem is to leverage pre-trained models. For the text modality, BERT \cite{devlin2018bert} is the most commonly used model and is well suited for feature extraction. Recently, speech-based pre-trained models have also emerged, and the most state-of-the-art of them is the wav2vec~2.0 \cite{baevski2020wav2vec}, which has shown promising performance in ASR \cite{zhang2020pushing}, speaker verification \cite{fan2020exploring} and emotion recognition \cite{zhao2022multi}.

Attention mechanisms enable models to learn which parts are important and which parts are irrelevant \cite{vaswani2017attention}. So it is well suited for handling sequential inputs, such as text and speech. Self-attention is an attention mechanism relating different positions of a single sequence to compute a representation of the sequence~\cite{vaswani2017attention}. Based on attention mechanisms, co-attention is a fusion method in which the representation of the sequence of a modality is computed by relating the sequence of the other modality~\cite{lu2019vilbert}. Several works have proposed to leverage pre-trained models of both speech and text modalities and attention mechanisms for multimodal emotion recognition and achieved promising results \cite{zhao2022multi,9746598,schneider19_interspeech,liu2019roberta}.

However, not all parts in the data sequences are related to emotion, and without the guidance of external knowledge, it is challenging for the attention to learn emotionally relevant parts accurately on its own \cite{bai2022enhancing}. To solve this problem, we propose to incorporate external emotion-related knowledge. More specifically, an emotion lexicon is leveraged with the attention mechanisms to improve its interpretability and robustness for the emotion recognition task. The emotion lexicon has shown its effectiveness in the previous methods. The work in~\cite{fu2021consk} used commonsense knowledge to build a graph and retrieved emotion-related information for the nodes in it using an emotion lexicon. Some other works proposed to integrate an emotion lexicon with contextual information for emotion recognition in conversations~\cite{9706271}.
However, there is no current work that combines the emotion lexicon with attention mechanisms. In this paper, we propose a Bayesian framework \cite{fan2020bayesian} with an emotion knowledge-related attention map as the prior distribution. The posterior distribution of the attention weights can then be learned. The reason that a Bayesian attention module (BAM) is introduced instead of simply adding the knowledge-related attention map on the original one is that the knowledge we introduce is not collected on the dataset we use, which may introduce bias. If a rather "hard" way is used for incorporating the knowledge, it may affect the learning of the attention model. The approach we propose allows the introduction of prior knowledge while achieving an integrated training of the attention weights, thus preventing this problem. We also show that the combination with the late fusion model further improves the performance.

We summarize our major contributions as follows:
\vspace{-0.3cm}

\begin{itemize}
\item We introduce emotion-related knowledge in the attention model to help the model focus more on relevant information for emotion recognition.
\vspace{-0.3cm}
\item To effectively incorporate knowledge, we further propose to use BAM, which not only brings randomness in the model and thus can help in modeling complex dependencies but also incorporates knowledge as prior in a "soft" way. We also show that it's complementary to the late fusion model.
\vspace{-0.3cm}
\item We evaluate the proposed models on the popular IEMOCAP dataset \cite{busso2008iemocap}. Experimental results show that they can outperform existing multimodal approaches using both speech and text modalities.
\vspace{-0.2cm}
\end{itemize}

\section{Proposed Methods}
\begin{figure}[t]
  \centering
  \includegraphics[width=8cm]{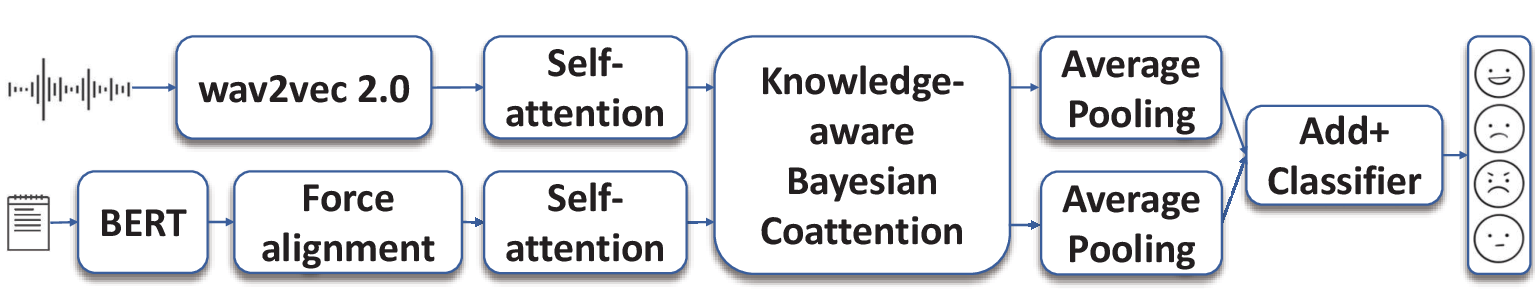}
  \vspace{-0.2cm}
  \caption{The proposed model. Word-level force alignment is applied for using a word-level emotion lexicon. Knowledge-aware Bayesian co-attention is used to fuse two modalities and also inject knowledge.}
  \label{1}
 \vspace{-0.5cm}
\end{figure}

\vspace{-0.2cm}
\label{sec:format}
The proposed model is shown in Figure~\ref{1}. First, we use BERT to extract text embeddings and wav2vec~2.0 to extract frame-level speech embeddings. Then we use word-level force alignment to get word-level speech embeddings in Section~\ref{fa}. Next, we apply self-attention to both embeddings to process input sequences and soften knowledge in Section~\ref{sf}. We introduce the general co-attention module in Section~\ref{gc} and the proposed knowledge-aware co-attention module to fuse two modalities in Section~\ref{ba}. Average pooling layers and linear layers are then used to output the probability distribution for emotion classification.
\vspace{-0.4cm}
\subsection{Word-level Force Alignment}
\vspace{-0.1cm}
\label{fa}
The knowledge we adopt is of word-level, and we want to introduce knowledge for both text and speech modalities, so we use the word-level force alignment to cope with frame-level speech embeddings:
\vspace{-0.2cm}
\begin{equation}
{\bf{u}}_k^{\rm{W}} = \frac{{\sum\nolimits_{f = {s_k}}^{{e_k}} {{{\bf{u}}_f}} }}{{{e_k} - {s_k}}},
\label{fali}
\vspace{-0.3cm}
\end{equation}
where ${\bf{u}}_k^{\rm{W}}$ is the $k^{\rm{th}}$ segmentation of the word-level speech embedding ${{\bf{U}}^{\rm{W}}}$, ${{\bf{u}}_f}$ is the $f^{\rm{th}}$ frame of the frame-level speech embedding, $s$ is the starting frame, $e$ is the end frame.
Segment-level speech embeddings like word-level speech embeddings are closely related to the prosody, and prosody can convey characteristics of the utterance like emotional state because it contains the information of the cadence of speech signals \cite{zhao2022multi}. 

\vspace{-0.3cm}

\subsection{Self-attention Module}
\vspace{-0.1cm}
\label{sf}
Self-attention is an attention mechanism relating different positions of a single sequence to compute a representation of the sequence \cite{vaswani2017attention}. Without loss of generality, the computation of one-head attention is given here:
\vspace{-0.1cm}
\begin{equation}
\{ {\bf{Q_s}},{\bf{K_s}},{\bf{V_s}}\}  = {\bf{W}}_{\{ 1,2,3\} }^i{{\bf{U}}^i} + b_{\{ 1,2,3\} }^i ,
\label{q}
\end{equation}
where ${{\bf{U}}^i}$ is the embedding for modality $i \in \{ {\rm{W,T}}\}$, $\rm{W}$ is for word-level speech embeddings and $\rm{T}$ is for text embeddings. $\bf{W}$ is the weight and $b$ is the bias. In the self-attention module, $i$ in Equation~\ref{q} are the same for $\bf{Q_s}$, $\bf{K_s}$ and $\bf{V_s}$. Thus, we can obtain the processed embeddings as:
\vspace{-0.2cm}
\begin{equation}
{\bf{M}} = {\rm{softmax}}\left( {\frac{{{\bf{Q_s}}{{\bf{K_s}}^ \top }}}{{\sqrt d }}} \right)
\label{attmap}
\vspace{-0.2cm}
\end{equation}
\vspace{-0.1cm}
\begin{equation}
{\rm{Attention}} = {\bf{MV}},
\label{attention}
\vspace{-0.1cm}
\end{equation}

where $d$ represents the dimension of the embeddings. Multi-head attention performs this process multiple times to learn information from various representation subspaces. We also use the self-attention module to soften the knowledge from the NRC VAD lexicon~\cite{mohammad2018obtaining}, which is the emotion lexicon we use. Because the knowledge is relatively sparse, without this step, the model will have difficulty learning useful information from the knowledge. The process of softening the knowledge is as follows: Given a sequence, we first look up the lexicon to find words that appeared in the lexicon. The VAD values we use range from -1 to 1. Then we compute the $\rm{L}^2$-norm of each word to get its intensity:
${i_k} = \sqrt[2]{{v_k^2 + a_k^2 + d_k^2}}$, 
where $i_k$ is the $k^{\rm{th}}$ of vector $\bf{i}$ which is the intensity for the sequence and $v$, $a$, $d$ are for valence, arousal, dominance respectively. Intensity is assigned to 0 for words that do not appear in the lexicon. Now we get the knowledge $\bf{i}$ for the sequence, but it faces the problem of sparsity. So we dot product the $k^{\rm{th}}$ row of the attention map in Equation \ref{attmap} with the $\bf{i}$ to get the softening intensity $i_k^{soft}$ which is the $k^{\rm{th}}$ of vector ${{\bf{i}}^{soft}}$:
\vspace{-0.2cm}
\begin{equation}
i_k^{soft} = {{\bf{M}}_k}{\bf{i}}.
\label{softknow}
\vspace{-0.3cm}
\end{equation}

\begin{figure}[t]
  \centering
  \includegraphics[width=8cm]{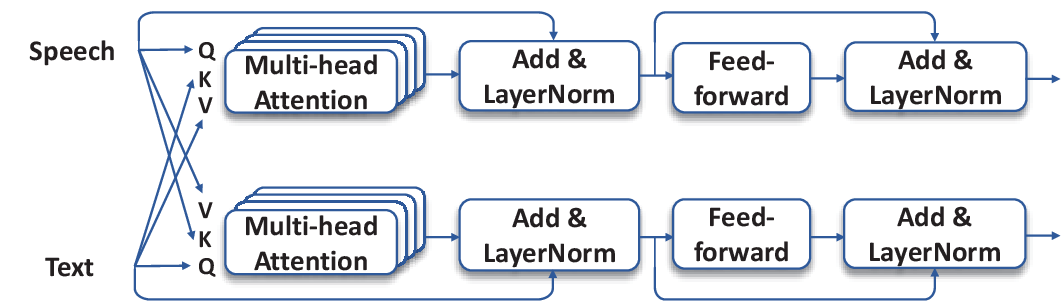}
  \vspace{-0.2cm}
  \caption{Co-attention module. Multi-head attention is used to fuse two modalities.}
  \label{2}
  \vspace{-0.3cm}
\end{figure}

\begin{figure}[t]
  \centering
  \includegraphics[width=7cm]{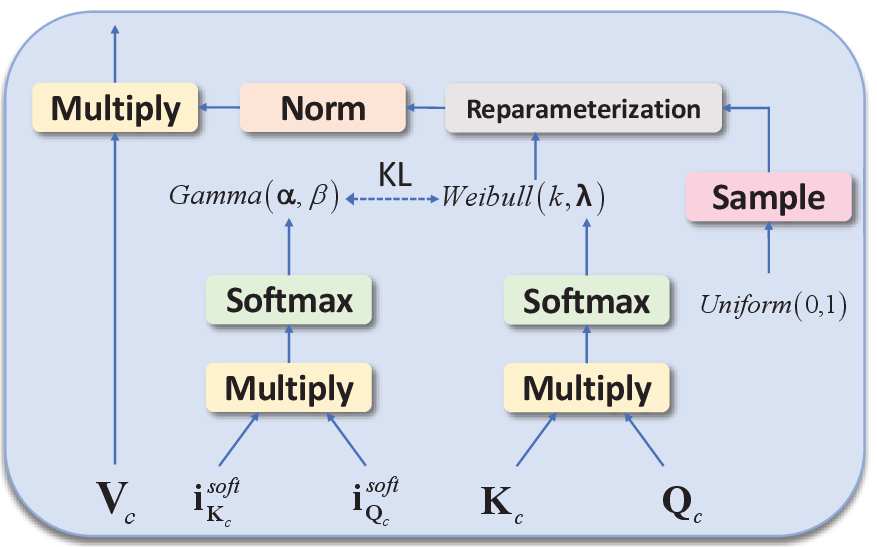}
  \caption{Knowledge aware Bayesian attention during training. ${\bf{i}}_{{\bf{Q_c}}/{\bf{K_c}}}^{soft}$ - intensity for the sequence which computes $\bf{Q_c/K_c}$ and is used to compute prior distribution. Reparameterization is used for the backpropagation of gradients. }
  \label{3}
  \vspace{-0.5cm}
\end{figure}
\vspace{-0.3cm}
\subsection{Co-attention Module}
\vspace{-0.1cm}
\label{gc}
The co-attention module \cite{lu2019vilbert} is depicted in Figure~\ref{2}. There are two branches in the co-attention module, each of which has the same structure as the Transformer encoder \cite{vaswani2017attention}. First embeddings from two modalities are sent to multi-head attention respectively. The computation of $\bf{Q_c}$, $\bf{K_c}$ and $\bf{V_c}$ is the same as Equation \ref{q}, but here each branch has different modalities as $\bf{Q_c}$ or $\bf{K_c}$,$\bf{V_c}$. Then attention map is computed using Equation \ref{attmap}. The output of multi-head attention is computed using Equation \ref{attention} and then added with its input which is used to compute its $\bf{Q_c}$. Layernorm layers are added to help the model converge \cite{ba2016layer}. The feed-forward layer is then applied, and it consists of two linear layers with a nonlinear function. The output of the feed-forward layer is added with its input and then Layernorm is applied. Co-attention module can be applied several times. The output is the mean of the two branches of the final co-attention module.
\vspace{-0.4cm}
\subsection{Knowledge-aware Bayesian Co-attention Module}
\vspace{-0.1cm}
\label{ba}
The proposed Knowledge-aware Bayesian Co-attention Module has the same structure as the co-attention module in Section~\ref{gc} except the computation of the attention map in Equation~\ref{attmap} which is shown in Figure~\ref{3}. In the BAM \cite{fan2020bayesian}, attention weights $\bf{W}$ are not deterministic but data-dependent local random variables from a variational distribution. It was shown that this variational distribution can approximate the posterior of attention weights $\bf{W}$ using reparameterizable attention distributions and also a prior as regularization. Following \cite{fan2020bayesian}, we use a Weibull distribution as the variational distribution and Gamma distribution as the prior distribution, then the regularization KL divergence can be computed as:
\vspace{-0.2cm}
\begin{equation}
\begin{split}
{\rm{KL}}\left( {{\rm{Weibull}}\left( {k,\lambda } \right)||{\rm{Gamma}}\left( {\alpha ,\beta } \right)} \right) =  \frac{{\gamma \alpha }}{k} - \alpha \log \lambda  \\+ \log k + \beta \lambda \Gamma \left( {1 + \frac{1}{k}} \right) - \gamma  - 1 - \alpha \log \beta  + \log \Gamma \left( \alpha  \right),
\label{KL}
\end{split}
\end{equation}
where $k$, $\lambda$ are the parameters of Weibull distribution, $\alpha$, $\beta$ are the parameters of Gamma distribution, $\gamma$ is the Euler's constant, and $\Gamma$ is the gamma function. The original attention map of the co-attention module computed using Equation~\ref{attmap} is used to compute $\lambda$ of Weibull distribution. The sample of the Weibull distribution after normalization is the attention map of the BAM during training. During inference, the posterior expectations are used to obtain the point estimates~\cite{srivastava2014dropout}. However, sampling directly from the distribution will fail backpropagation of gradients. Following \cite{fan2020bayesian}, the Weibull distribution is a reparameterizable distribution, so this problem can be solved by sampling from the standard uniform distribution $\varepsilon  \sim {\rm{Uniform}}\left( {0,1} \right)$. Then the sample from Weibull distribution is equivalent to drawing: ${\bf{S}}: = \lambda {\left( { - \log \left( {1 - \varepsilon } \right)} \right)^{1/k}}
\nonumber$, where $\lambda$, $k$ are the parameter of Weibull distribution. The sample ${\bf{S}}$ is multiplied with $\bf{V_c}$ to get the output of the knowledge-aware Bayesian attention, similar to Equation \ref{attention}.

Unlike \cite{fan2020bayesian}, which uses $\bf{K_c}$ to compute parameters of the prior distribution in Equation \ref{KL}, we use the knowledge vector ${\bf{i}}_{\bf{Q_c}}^{soft}$, ${\bf{i}}_{\bf{K_c}}^{soft}$ in Equation \ref{softknow} to compute them where ${\bf{i}}_{{\bf{Q_c}}/{\bf{K_c}}}^{soft}$ is the intensity for the sequence which computes $\bf{Q_c/K_c}$:
${\bf{P}} = {\rm{softmax}}\left( {{\bf{i}}{{_{\bf{Q_c}}^{soft}}^ \top }{\bf{i}}_{\bf{K_c}}^{soft}} \right)$.
We use $\bf{P}$ to compute $\alpha$ of the Gamma distribution in Equation \ref{KL}. In this way, prior knowledge can be introduced while still leaving the attention weights to be learned by the model itself, which alleviates the problem of bias in the emotion lexicon. In this way the $k^{\rm{th}}$ of the intensity $i_{{\bf{Q_c}},k}^{soft}$ works like temperature for ${\bf{i}}_{\bf{K_c}}^{soft}$ in every row. If $i_{{\bf{Q_c}},k}^{soft}$ is high, then this word is highly related to emotion, and the temperature will strengthen its connection with other emotional words. If $i_{{\bf{Q_c}},k}^{soft}$ is low, a preposition for example, then this word may not be related to emotion, and the temperature will weaken its connection with other emotional words. The $\beta$ of the Gamma distribution and the $k$ of the Weibull distribution remain as hyperparameters.
\vspace{-0.3cm}

\section{Experiments and Discussion}
\vspace{-0.3cm}
\subsection{Datasets}
\vspace{-0.2cm}
The IEMOCAP dataset is an acted, multimodal and multispeaker dataset with five sessions \cite{busso2008iemocap}. We use 4 emotional classes: anger, happiness, sadness, and neutral. The number of utterances representing them was 1103, 1636, 1084, and 1708 respectively. Alignment information of IEMOCAP is provided in their datasets. Other datasets can also get alignment information using force alignment tools such as Prosody Aligner \cite{gorman2011prosodylab} and Montreal Forced Aligner \cite{mcauliffe2017montreal}. We use the NRC VAD lexicon \cite{mohammad2018obtaining} as the emotion lexicon. It includes a list of more than 20,000 English words and their valence (positiveness-negativeness), arousal (active-passive), and dominance (dominant-submissive) scores.  
\vspace{-0.5cm}

\subsection{Settings and Metrics}
\vspace{-0.2cm}
Cross-entropy loss plus KL loss in Equation~\ref{KL} was used as the loss function, and an Adam optimizer \cite{kingma2014adam} was applied using a learning rate of 1e-4. The models were trained using a batch size of 32 and early stopping was applied with a patience of 6. Dropout \cite{srivastava2014dropout} was applied with a probability of 0.1 after every feed-forward layer except the output layer to prevent overfitting. We use wav2vec~2.0-base and BERT-base un-cased models, both of which have 768-dimensional embeddings. Following \cite{baevski2020wav2vec}, we extract embeddings from all 12 transformer encoder layers in wav2vec 2.0, and we also apply the weighted average for different layers to get the final word-level speech embeddings. Similarly, we also apply the weighted average for the text modality. All speech samples are normalized by global normalization which is a frequently used setting for this dataset. $k$ of Weibull distribution was set to 1 and $\beta$ of Gamma distribution was set to 10. Five-fold cross-validation was repeated 5 times and the results were the average of them. A frame length of 400 is used. We observed that more layers of the self-attention module and co-attention module didn't bring improvement in all models, so they were set to 1. Unweighted accuracy (UA) and weighted accuracy (WA) were used as our evaluation metrics.

\vspace{-0.5cm}

\subsection{Results of the Single Modality}
\vspace{-0.2cm}
The first part of Table \ref{kw} shows the results of the single modality. They have the same structure as one branch of Figure \ref{1} except that there is no co-attention module. Word-level speech embeddings are computed from frame-level speech embeddings using Equation \ref{fali}. Text modality has the best performance. This is because in IEMOCAP there are not many complex situations where speech is more effective, such as sarcasm. We can also observe that using the word-level alignment information still has similar results compared to the complete frame-level speech embeddings, but with a much smaller size.
\vspace{-0.5cm}

\begin{table}[]
\caption{Results of the single modality and multimodality. T - text, F - frame-level speech, W - word-level speech, Self-att - self-attention, Co-att - co-attention, BAM - Bayesian attention module}
\vspace{0.1cm}
\resizebox{80mm}{!}{
\label{kw}
\begin{tabular}{cccc}
\toprule
\multicolumn{1}{l}{\textbf{Modalities}} & \textbf{Models}                     & \multicolumn{1}{l}{\textbf{UA}} & \multicolumn{1}{l}{\textbf{WA}} \\ \midrule
T                                       & Self-att                     & 66.6\%                          & 65.2\%                          \\
F                                       & Self-att                     & 63.8\%                          & 62.2\%                          \\
W                                       & Self-att                      & 63.4\%                          & 61.7\%                          \\ \midrule
T+F                                     & Co-att                       & 74.8\%                          & 73.5\%                          \\
T+W                                     & Co-att                      & 74.9\%                          & 73.6\%                          \\
T+W                                     & Co-att+knowledge              & 74.6\%                          & 73.3\%                          \\
T+W                                     & Co-att+BAM                     & 74.8\%                          & 73.7\%                          \\

\textbf{T+W}                            & \textbf{Co-att+BAM+knowledge} & \textbf{75.6\%}                 & \textbf{74.0\%}     \\ \bottomrule           
\end{tabular}}
\vspace{-0.5cm}
\end{table}

\subsection{Results of the Multimodality}
\vspace{-0.2cm}
The second part of Table~\ref{kw} shows the results of multimodality. Compared to the results of the single modality, we can see that the multimodality can yield good performance improvements which are consistent with that observed in \cite{zhao2022multi,9746598}. It can also be seen that the word-level speech embedding-based co-attention fusion gives a similar performance as the frame-level one. Next, the results of word-level co-attention with a knowledge map directly added to the original attention map (Co-att+knowledge) are shown. It's a rather "hard" way which is shown to degrade the performance slightly. We show the results of word-level co-attention with BAM (Co-att+BAM) next, in which we use the prior \cite{fan2020bayesian} computed using $\bf{K_s}$ in Equation~\ref{q}. We can observe that incorporating the BAM alone without knowledge cannot improve the model performance. This may be because it's challenging to train BAM to fuse two modalities for multimodal emotion recognition without external guidance.  Finally, with the BAM and prior estimated using the emotion-related knowledge (Co-att+BAM+knowledge), it yields the best results among all the co-attention-based models, which validates the effectiveness of the proposed knowledge-enhance BAM model.
\vspace{-0.5cm}
\subsection{Results of the Combination with Late Fusion}
\vspace{-0.2cm}
In \cite{zhao2022multi}, they show that the co-attention model can combine with the late fusion model simply with score fusion to boost the performance. Table~\ref{scorefusion} shows that this combination can also give performance gains for our knowledge-aware co-attention. The late fusion model that fuses text and word-level speech embeddings has the same structure as that in \cite{zhao2022multi}. The late fusion model that fuses text and frame-level speech embeddings has a similar performance and is omitted here. We can see that late fusion with the knowledge-aware Bayesian co-attention model has the best performance, yielding 77\% UA and 75.5\% WA, which demonstrates the complementarity of the late fusion and knowledge-aware Bayesian co-attention.
\vspace{-0.5cm}

\subsection{Comparison with Existing Methods}
\vspace{-0.2cm}
We compare the proposed method with state-of-the-art multimodal emotion recognition methods in Table~\ref{sotamethod}. For a fair comparison, all the experiments use the text and speech data from IEMOCAP, 5-fold cross-validation, and four classes. It shows that our proposed model achieves the best performance and surpasses other methods by at least 0.7\% UA.
\vspace{-0.5cm}

\begin{table}[]
\caption{Results of the combination with late fusion}
\vspace{0.1cm}
\begin{center}
\resizebox{80mm}{!}{
\label{scorefusion}
\begin{tabular}{ccc}
\toprule
\textbf{Models}                                               & \textbf{UA}     & \textbf{WA}     \\ \midrule
Late fusion w/o Co-att                                      & 75.7\%          & 74.2\%          \\ 
Late fusion+Co-att        & 76.3\%          & 75.2\%          \\ 
\textbf{Late fusion+Co-att+BAM+knowledge}  & \textbf{77.0}\%          & \textbf{75.5\%}          \\ 

 \bottomrule
\end{tabular}}
\end{center}
\vspace{-0.7cm}
\end{table}

\begin{table}[]

\caption{Comparison with existing methods}

\begin{center}
\label{sotamethod}
\begin{tabular}{ccc}
\toprule
\textbf{Models}                                               & \textbf{UA}     & \textbf{WA}     \\ \midrule
Chen et al. \cite{chen20b_interspeech}, 2020                             & 72.1\%          & 71.1\%          \\ 
Makiuchi et al. \cite{makiuchi2021multimodal}, 2021                     &
73.0\%          & 73.5\% \\
Kumar et al. \cite{kumar2021towards}, 2021                              &
75.0\%          & 71.7\% \\
Chen et al. \cite{9746598}, 2022         & 75.3\%          & 74.3\%          \\ 
Zhao et al. \cite{zhao2022multi}, 2022 & 76.3\%          & -          \\ 
\textbf{Ours}  & \textbf{77.0}\%          & \textbf{75.5\%}          \\ 
\bottomrule
\end{tabular}
\end{center}
\vspace{-0.8cm}
\end{table}

\section{Conclusion}
\vspace{-0.3cm}
In this paper, we have proposed to incorporate emotion-related knowledge in Bayesian co-attention modules for multimodal emotion recognition. Experimental results have shown that the proposed Bayesian co-attention model can outperform the baseline multimodal emotion recognition methods and achieves an accuracy of 77.0\% UA and 75.5\% WA on the 5-fold CV on IEMOCAP.

\bibliographystyle{IEEEbib}
\bibliography{strings,refs}

\end{document}